\documentclass[graybox]{svmult}

\usepackage{mathptmx}       % selects Times Roman as basic font
\usepackage{helvet}         % selects Helvetica as sans-serif font
\usepackage{courier}        % selects Courier as typewriter font
\usepackage{type1cm}        % activate if the above 3 fonts are
\usepackage{makeidx}         % allows index generation
\usepackage{graphicx}        % standard LaTeX graphics tool
\usepackage{multicol}        % used for the two-column index
\usepackage[bottom]{footmisc}% places footnotes at page bottom
\usepackage{amsfonts}

\makeindex             % used for the subject index
\begin{document}

\title*{Learning to Communicate with Reinforcement Learning for an Adaptive Traffic Control System}
\titlerunning{Learning to Communicate with Reinforcement Learning for an ATCS}
% Use \titlerunning{Short Title} for an abbreviated version of
% your contribution title if the original one is too long
\author{Simon Vanneste, Gauthier de Borrekens, Stig Bosmans, Astrid Vanneste, Kevin Mets, Siegfried Mercelis, Steven Latré, Peter Hellinckx}
\authorrunning{Simon Vanneste et al.}
% Use \authorrunning{Short Title} for an abbreviated version of
% your contribution title if the original one is too long
\institute{
    Simon Vanneste\textsuperscript{\rm 1},
    Gauthier de Borrekens\textsuperscript{\rm 1},
    Stig Bosmans\textsuperscript{\rm 1},
    Astrid Vanneste\textsuperscript{\rm 1},
    Kevin Mets\textsuperscript{\rm 2},
    Siegfried Mercelis\textsuperscript{\rm 1},
    Steven Latré\textsuperscript{\rm 2},
    Peter Hellinckx\textsuperscript{\rm 1}
    \at {\textsuperscript{\rm 1}
    University of Antwerp - imec
    IDLab - Faculty of Applied Engineering
    Sint-Pietersvliet 7, 2000 Antwerp, Belgium
    \\\textsuperscript{\rm 2}
    University of Antwerp - imec
    IDLab - Department of Computer Science
    Sint-Pietersvliet 7, 2000 Antwerp, Belgium}\\
    \email{
simon.vanneste@uantwerpen.be
gauthier.deborrekens@uantwerpen.be,
stig.bosmans@uantwerpen.be,
astrid.vanneste@uantwerpen.be,
kevin.mets@uantwerpen.be,
siegfried.mercelis@uantwerpen.be,
steven.latre@uantwerpen.be,
peter.hellinckx@uantwerpen.be
}}

\thispagestyle{empty} 
\section*{Copyright Notice}
This is a preprint of the following chapter: Simon Vanneste, Gauthier de Borrekens, Stig Bosmans, Astrid Vanneste, Kevin Mets, Siegfried Mercelis, Steven Latré, Peter Hellinckx, Learning to Communicate with Reinforcement Learning for an Adaptive Traffic Control System, published in Advances on P2P, Parallel, Grid, Cloud and Internet Computing. 3PGCIC 2021. Lecture Notes in Networks and Systems, vol 343., edited by Leonard Barolli, 2021, Springer reproduced with permission of Springer Nature Switzerland AG. The final authenticated version is available online at: http://dx.doi.org/10.1007/978-3-030-89899-1\_21.

{
	\let\clearpage\relax
	\maketitle
}

%
% Use the package "url.sty" to avoid
% problems with special characters
% used in your e-mail or web address
%

\abstract*{
Recent work in multi-agent reinforcement learning has investigated inter agent communication which is learned simultaneously with the action policy in order to improve the team reward.
In this paper, we investigate independent Q-learning (IQL) without communication and differentiable inter-agent learning (DIAL) with learned communication on an adaptive traffic control system (ATCS).
In real world ATCS, it is impossible to present the full state of the environment to every agent so in our simulation, the individual agents will only have a limited observation of the full state of the environment.
The ATCS will be simulated using the Simulation of Urban MObility (SUMO) traffic simulator in which two connected intersections are simulated.
Every intersection is controlled by an agent which has the ability to change the direction of the traffic flow.
Our results show that a DIAL agent outperforms an independent Q-learner on both training time and on maximum achieved reward as it is able to share relevant information with the other agents.
}

\abstract{
Recent work in multi-agent reinforcement learning has investigated inter agent communication which is learned simultaneously with the action policy in order to improve the team reward.
In this paper, we investigate independent Q-learning (IQL) without communication and differentiable inter-agent learning (DIAL) with learned communication on an adaptive traffic control system (ATCS).
In real world ATCS, it is impossible to present the full state of the environment to every agent so in our simulation, the individual agents will only have a limited observation of the full state of the environment.
The ATCS will be simulated using the Simulation of Urban MObility (SUMO) traffic simulator in which two connected intersections are simulated.
Every intersection is controlled by an agent which has the ability to change the direction of the traffic flow.
Our results show that a DIAL agent outperforms an independent Q-learner on both training time and on maximum achieved reward as it is able to share relevant information with the other agents.
}

%%%%%%%%%%%%%%%%%%%%%%%%%%%%%%%%%%%%%%%%%%%%%%%%%%%%%%%%%%%%%%%%%%%%%%%%%%%%%%%%%%%%%%%%%%%%%%%%%%%%%%%%%%%%%%%
% Introduction
%%%%%%%%%%%%%%%%%%%%%%%%%%%%%%%%%%%%%%%%%%%%%%%%%%%%%%%%%%%%%%%%%%%%%%%%%%%%%%%%%%%%%%%%%%%%%%%%%%%%%%%%%%%%%%%
\section{Introduction}
\label{sec:introduction}
% problems with traffic (CO2 and gdp europe)
Traffic congestion is a worldwide problem which has a big environmental and economical impact. It has been estimated that in the EU, most traffic congestion's occur around urban areas and that these congestion's cost nearly 1\% of the EU's GDP \cite{european2011roadmap}. In this work, we will investigate traffic light control in order to improve congestion in urban area.
% RL could be a tool to improve this
Reinforcement learning (RL) is a machine learning method that searches for a policy which will maximize the future expected reward. RL has been demonstrated to successfully learn a policy in a wide area of environments like the Atari environments \cite{mnih2013playing}. RL has also successfully been applied to intelligently managing traffic lights \cite{ThorpeSARSA}\cite{zheng2019diagnosing}.
% Multi agent
Single agent RL implementations have shown effective in simplified traffic environments, but suffer in more complex and realistic scenarios. To address this, Multi-Agent Reinforcement Learning (MARL) has been proposed, in which every traffic light is controlled by a single agent. These independent agents still have access to the full state of the environment.
% Dec-MDP + communication
In this work, we will investigate a MARL system with agents that do not have access to the full state but only to their local observation for the Adaptive Traffic Control System (ATCS) environment. In order to overcome this partial observability, the agents have the ability to communicate with each other in order to share information about their observation with the other agents. The communication protocol is not predefined, but are fully enveloped in the reinforcement learning process, meaning the agents will learn how to communicate during the reinforcement learning process. If the agents do not discover the benefit of communication, no emergence will take place.
% We use DIAL
Independent Q-learning and Differential Inter Agent Learning (DIAL) \cite{foerster2016learning} are trained on a simplified traffic environment, simulated with the open-source simulator Simulation of Urban Mobility (SUMO) \cite{SUMO2018}. The FLOW framework is used as an interface between the RL algorithm and the simulated traffic environment \cite{wu2017flow}.
% Paper structure
The paper is structured as followed.
In Section \ref{sec:related_work}, we will discuss the related work. Section \ref{sec:method} describes the used reinforcement learning methods. The environment used to train these reinforcement learning agent is discussed in Section \ref{sec:Environment}. Next, we describe the results in Section \ref{sec:results}. Finally, we present our conclusion and future work in Section \ref{sec:conclusion}.

%%%%%%%%%%%%%%%%%%%%%%%%%%%%%%%%%%%%%%%%%%%%%%%%%%%%%%%%%%%%%%%%%%%%%%%%%%%%%%%%%%%%%%%%%%%%%%%%%%%%%%%%%%%%%%%
% Related Work
%%%%%%%%%%%%%%%%%%%%%%%%%%%%%%%%%%%%%%%%%%%%%%%%%%%%%%%%%%%%%%%%%%%%%%%%%%%%%%%%%%%%%%%%%%%%%%%%%%%%%%%%%%%%%%%
\section{Related Work}
\label{sec:related_work}
This section will discuss the related work in the domain of MARL for the ATCS application and the related work in communication learning using MARL.
% Multiagent reinforcement learning for integrated network of adaptive traffic signal controllers (MARLIN-ATSC): methodology and large-scale application on downtown Toronto
% https://ieeexplore.ieee.org/stamp/stamp.jsp?arnumber=6502719&casa_token=_fgLwIa7EhcAAAAA:cbSUcf9rlRvzrEWlaQL-qkuVyEnw1cEsVOH89Fm2W2geEDwPTIiWNDkM4X20-byscqHFV8OFJraUyA&tag=1
El-Tantawy et al. \cite{el2013multiagent} present a MARL ATCS called MARLIN-ATCS which is an independent Q-learning method that has a coordination system with neighbouring agents. This method is shown to outperform an independent Q-learner and real world controllers on the simulated network of Downtown Toronto.
% Coordinated deep reinforcement learners for traffic light control \cite{van2016coordinated}
Van der Pol et al. \cite{van2016coordinated} presented a multi-agent reinforcement learning system in which the global Q-function is learned as a linear combination of local subproblems. The subproblems are trained on the basic source problems containing two agents. Next, the global Q-function is used in the max-plus coordination \cite{kok2005using} algorithm to optimize the actions of the agents. This methods allows the agent to learn an action policy while circumventing the non-stationarity problem and the high training costs of MARL.
% Cooperative deep reinforcement learning for large-scale traffic grid signal control \cite{tan2019cooperative}
% https://ieeexplore.ieee.org/stamp/stamp.jsp?arnumber=8676356&casa_token=7awpf1nKsWAAAAAA:vE_zMhw05CBvhOWpt36yBrK9b87P8dYHTYbfjkaIWENL7C35vnt1dwygr4tzzX5C5Z6KM7JHWLTSYg
Tan et al. \cite{tan2019cooperative} presented a hierarchical multi-agent reinforcement learning system in which several regional controllers are combined with a centralized global agent. The regional controllers will all present a combination of actions which is further optimized by the centralized global agents.
% Our work
These methods focus on the challenges of large scale learning in an ATSC system. In this work we focus on the challenge of learning in a decentralized Markov decision process (Dec-MDP) \cite{oliehoek2016concise} in which agents learn to communicate in order to share information about their local observation.

The MARL domain is an active research domain in which several MARL communication learning methods have been developed.
% CommNet
Sukhbaatar  et  al. \cite{sukhbaatar2016learning} presented the CommNet RL method that will encode the communication information into the hidden state. This hidden state will be shared with the other agents and combined with their hidden state into the final hidden state which will be used by the agent to generate an action. This communication is learned end-to-end by allowing the gradients to flow through the agents based on the loss of the receiving agents.
% RIAL and DIAL
Foerster  et  al. \cite{foerster2016learning} presented two methods to learn to communicate. Reinforced Inter-Agent Learning (RIAL) will use Q-learning to select both the action and the message to send to the other agents. Differentiable Inter-Agent Learning (DIAL) will train the communication end-to-end as discussed in section \ref{sec:dial}. Their experiments show that DIAL outperforms RIAL because of the direct feedback the receiving agent can give to the sending agent using the end-to-end training.
% value-decomposition network + centralized critic
DIAL has been extended by Vanneste et al. \cite{vanneste2019learning} to include a value-decomposition network \cite{sunehag2017value}. The value-decomposition network can be used to improve the lazy agent problem in which a low performing agent is rewarded for the performance of the other agents.
Another method to reduce the lazy agent problem while learning to communicate is by using a centralized critic as shown in the MADDPG method \cite{lowe2017multi} and the MACC method \cite{vanneste2020learning}.

%%%%%%%%%%%%%%%%%%%%%%%%%%%%%%%%%%%%%%%%%%%%%%%%%%%%%%%%%%%%%%%%%%%%%%%%%%%%%%%%%%%%%%%%%%%%%%%%%%%%%%%%%%%%%%%
% Method
%%%%%%%%%%%%%%%%%%%%%%%%%%%%%%%%%%%%%%%%%%%%%%%%%%%%%%%%%%%%%%%%%%%%%%%%%%%%%%%%%%%%%%%%%%%%%%%%%%%%%%%%%%%%%%%
\section{Method}
\label{sec:method}
In this section, independent Q-learning and Differentiable Inter-Agent Learning are described. These two methods will be used to train our agent for the ATCS.

\subsection{Independent Q-Learning}
In the traditional single agent RL setting, the agent will receive the state $s_t \in S$ at timestep $t$ and respond with an action $u_t \in U$. The environment will use this action to present the next state $s_{t+1}$ and reward $r_t$ to the agent which will be used to train the agent. Q-learning learns a Q-value for every state-action pair $Q(s,u)$ in which a higher Q-value represents a better action. The RL policy can be defined as $\pi(s) = argmax_u Q(s,u)$. Deep Q-learning will use a deep Q-network (DQN) with network parameters $\theta$ to represent this Q-function. This network will be trained by minimizing the loss function which is show in Equation \ref{eq:dqn_loss}. The discount factor $\gamma$ is a value between 0 and 1 and represents the balance between long term and short term goals.

\begin{equation}
\mathcal{L}_i(\theta_i) = \mathbb{E}_{s_t,u_t,r_t,s_{t+1}} \big[ (r_t + \gamma \max_{u_{t}} Q(s_{t+1},u_{t+1}, \theta_i) - Q(s_t,u_t, \theta_i))^2\big]
\label{eq:dqn_loss}
\end{equation}

% IQL
An extension to DQN is independent Q-learning (IQL)\cite{tan1993multi} which allow us to train agents in a cooperative multi-agent setting. In IQL, every agent $a$ will train an individual Q-network $Q^a(s_t, u_t^a; \theta^a)$ based on the global state $s_t$, the individual action of the agent $u_t^a$ and the team reward $r_t$. The agent receive the same team reward in order to encourage cooperative behaviour.
% no replay buffer due to the environment becoming non stationary as the other agents are learning

\subsection{Differentiable Inter-Agent Learning}
\label{sec:dial}
% TODO
% Centralized Learning, Decentralized Execution (CLDE)

IQL assumes that the full state $s_t$ is available to every agent. In real-world applications, agents often have only a partial observation $o_t$ of the environment. In this work, we will assume that our problem can be represented as a decentralized Markov decision process (Dec-MDP) \cite{oliehoek2016concise} which means that the combination of all the observations of the different agents again represent the full state. In order to overcome this partial observability, the agents will have the ability to communicate with each other. The communication will allow the agent to send relevant observation information to other agent in order to recreate the state as required by IQL.

\begin{figure}[h]
\centering
% \sidecaption
\includegraphics[scale=0.8]{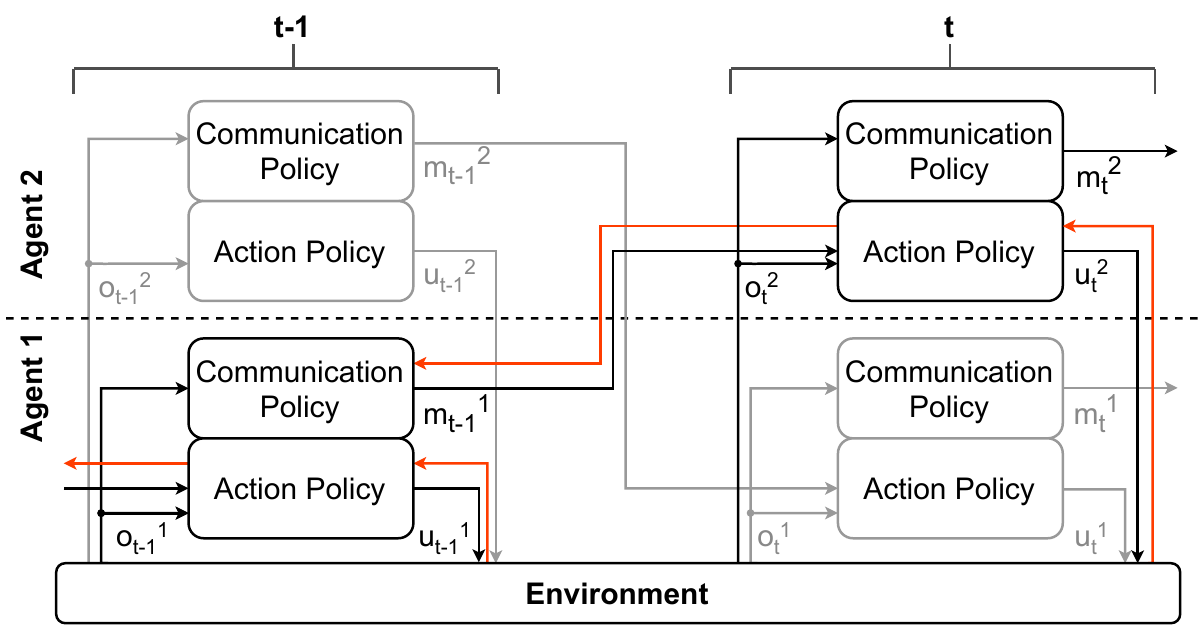}
\caption{DIAL with a separate action and communication policy.}
\label{fig:dial}       % Give a unique label
\end{figure}

The Differentiable Inter-Agent Learning\cite{foerster2016learning} (DIAL) method is an extension to the IQN in order to allow agent to learn an inter agent communication protocol. This communication protocol is not predefined but is learned simultaneously with the action policy. The action part of DIAL is identical to IQL apart from the fact that the action policy will also have access to the received messages. The DIAL method will learn the communication end-to-end by allowing the gradients to flow from the receiving agent back to the sending agents. These gradients are calculated in order to improve the IQN loss function of the independent Q-learner. DIAL can be used to learn continuous or discrete messages. In this work, we used continuous messages but this can be altered to discrete messages by including a discretise/regularise unit\cite{foerster2016learning}.
We have made a slight alteration to DIAL by splitting the action and communication policy without sharing any network parameters between these policies. This is possible because in our application, we are focusing on overcoming the partial observability of the Dec-MDP and no information is shared between the action and communication policy. This separation is however not possible when agents need to communicate strategic decisions with each other. The DIAL training process is illustrated in Figure \ref{fig:dial}.

%%%%%%%%%%%%%%%%%%%%%%%%%%%%%%%%%%%%%%%%%%%%%%%%%%%%%%%%%%%%%%%%%%%%%%%%%%%%%%%%%%%%%%%%%%%%%%%%%%%%%%%%%%%%%%%
% Environment
%%%%%%%%%%%%%%%%%%%%%%%%%%%%%%%%%%%%%%%%%%%%%%%%%%%%%%%%%%%%%%%%%%%%%%%%%%%%%%%%%%%%%%%%%%%%%%%%%%%%%%%%%%%%%%%
\section{Environment}
\label{sec:Environment}
\begin{figure}[h]
\centering
\includegraphics[scale=0.235]{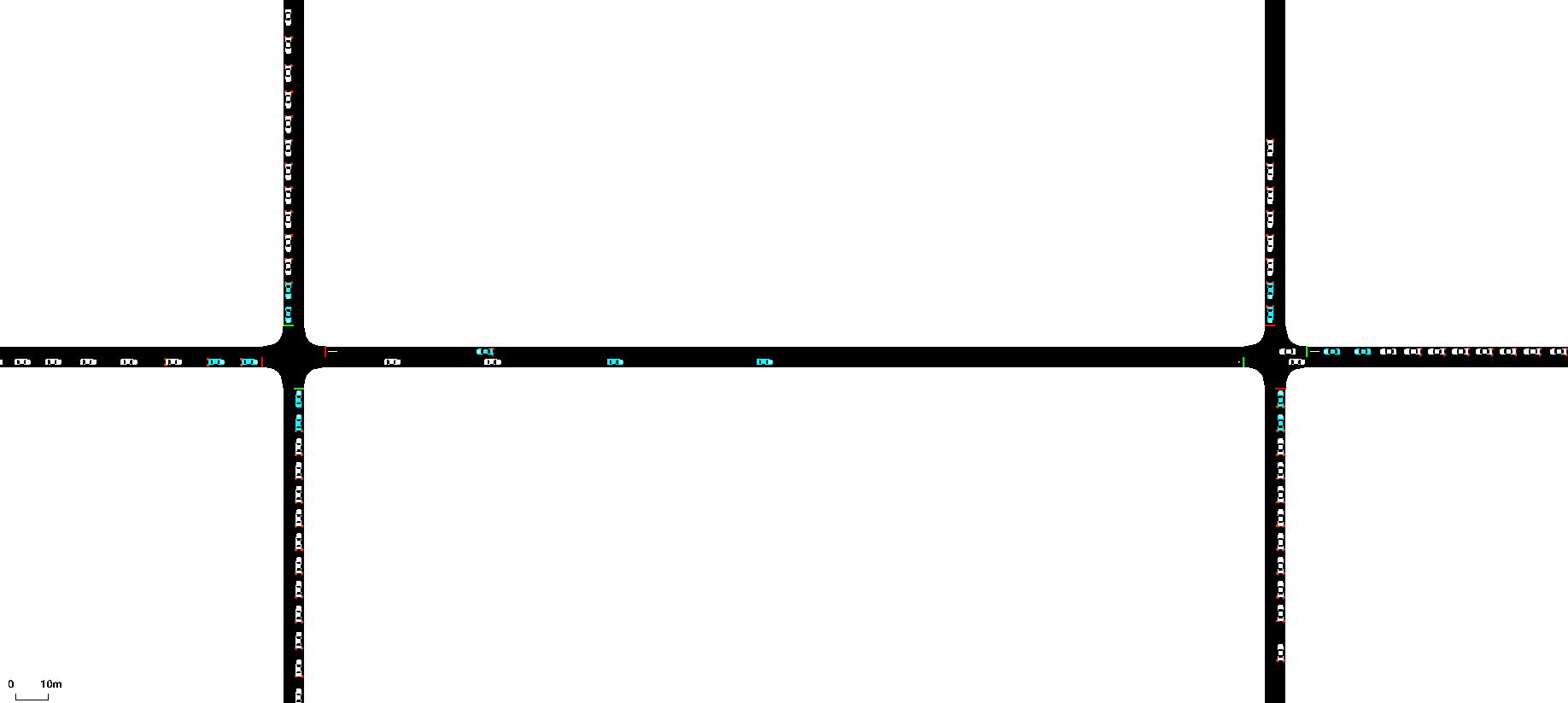}
\caption{A SUMO simulation with a network consisting of two intersections of four lanes, connected by a horizontal lane and currently being populated
with vehicles.}
\label{fig:SUMO}
\end{figure}

% SUMO
Reinforcement learning requires an environment to provide observations, execute the actions and evaluate the behaviour of an agent. To simulate the traffic, the Simulation of Urban MObility (SUMO) simulator \cite{SUMO2018} is used. SUMO allows for large-scale, continuous simulation with precise control over many factors.
% FLOW
The FLOW project \cite{wu2017flow} is a deep reinforcement learning framework that provides an interface for implementing and training RL algorithms on top of the SUMO simulator.
It consists of both a network and an environment.
% FLOW Network
The FLOW network contains all the information about the SUMO traffic simulation itself. It consists of a traffic layout defining the lanes and intersections. This layout is populated with vehicles and two traffic lights, each having a logic controller that defines their behaviour. For the vehicles, the built-in controller algorithm is used which will mimic human-like braking and acceleration behaviour.
For the traffic lights, we substitute the controller with an IQL or DIAL agent.
% FLOW Environment
The FLOW environment acts as the interface between the RL agent and the simulation. It defines the observation space that is returned to the agents, performs the agent actions on the environment and defines the reward function.
In Figure \ref{fig:SUMO}, our SUMO simulation is shown where the network consists of a grid of two intersections connected by a shared road. Both intersections contain traffic lights that are controlled by separate RL agents.

% Observation
The 26 values of the observation are shown in Table \ref{table:obs}. The first 12 values contain information on the speed, distance and position of the leading vehicle on each lane going toward the intersection. The edge number is used to identify which lanes the agents are observing, providing them information on their position in the environment. The following 12 values contain information on the traffic congestion, waiting queue and total vehicle waiting time of each of the four adjacent lanes. The final two bits indicate which direction the intersection is allowing through and if it is currently switching, indicated by a yellow light. Each agent also observes the communication messages which consist of 5 continuous values.

\begin{table}[h]
\centering
\begin{tabular}{|c c c c|}
\hline
0:3 & 4:7 & 8:11 & 12:15\\
Speed & distance to edge & edge number  & total vehicles\\
\hline
16:19 & 20:23 & 24 & 25\\
waiting vehicles & waiting time & direction & currently yellow\\
\hline
\end{tabular}
\caption{The 26 dimension observation space representing the current state of the environment.}
\label{table:obs}
\end{table}

The action consists of a single value that is used to switch directions of the traffic on the intersection.
The reward function aims to optimize global waiting times and is formulated as followed. Each vehicle contains an accumulated waiting time value that is incremented by one for each timestep its velocity is below 2 miles per hour. When the vehicle surpasses this velocity threshold, the total waiting time is decremented by 0.4 each timestep. The reward is the negative of the average of all accumulated waiting times at that timestep of both intersections. This reward was chosen in order to find a balance between waiting time and throughput on the intersections.

%%%%%%%%%%%%%%%%%%%%%%%%%%%%%%%%%%%%%%%%%%%%%%%%%%%%%%%%%%%%%%%%%%%%%%%%%%%%%%%%%%%%%%%%%%%%%%%%%%%%%%%%%%%%%%%
% Results
%%%%%%%%%%%%%%%%%%%%%%%%%%%%%%%%%%%%%%%%%%%%%%%%%%%%%%%%%%%%%%%%%%%%%%%%%%%%%%%%%%%%%%%%%%%%%%%%%%%%%%%%%%%%%%%
\section{Results}
\label{sec:results}
This section will evaluate both IQL and DIAL in the traffic light environment. In the first subsection, we will evaluate the training behaviour of IQL and DIAL. In the second subsection, the trained policies are evaluated on different levels of inflow. These experiments are trained using a DIAL implementation within the RLlib \cite{liang2018rllib} framework.
% Configuration Agents
In both experiments, the IQL and DIAL method have the same neural network architecture for the action policy. This action neural network consists of two layers of 256 neurons and a ReLu activation function. The final layer consists of a linear layer with and output size of 2. The hyperparameters for both methods are shown in Table \ref{table:hyperparameters}. The DIAL method has an additional communication neural network which consists of no hidden layers as this network will make an encoding of the observation into a messages with 5 dimensions. This network could be expanded in order to reduce the size of the message or to move processing from the receiver towards the sender. In this work, we chose a linear mapping between observations and message in order to make the messages more understandable.

\begin{table}[h]
\centering
\begin{tabular}{|c|c|}
\hline
Parameter & Value\\
\hline
Gamma & 0.99\\
Epsilon & 1\\
Epsilon Decay & 0.99995\\
Minimal Epsilon & 0.05\\
Learning rate & 0.0005\\
Optimizer & Adam\\
\hline
\end{tabular}
\caption{The IQL and DIAL hyperparameters.}
\label{table:hyperparameters}
\end{table}

\subsection{Training}
\begin{figure}[h]
\sidecaption
\includegraphics[scale=.47]{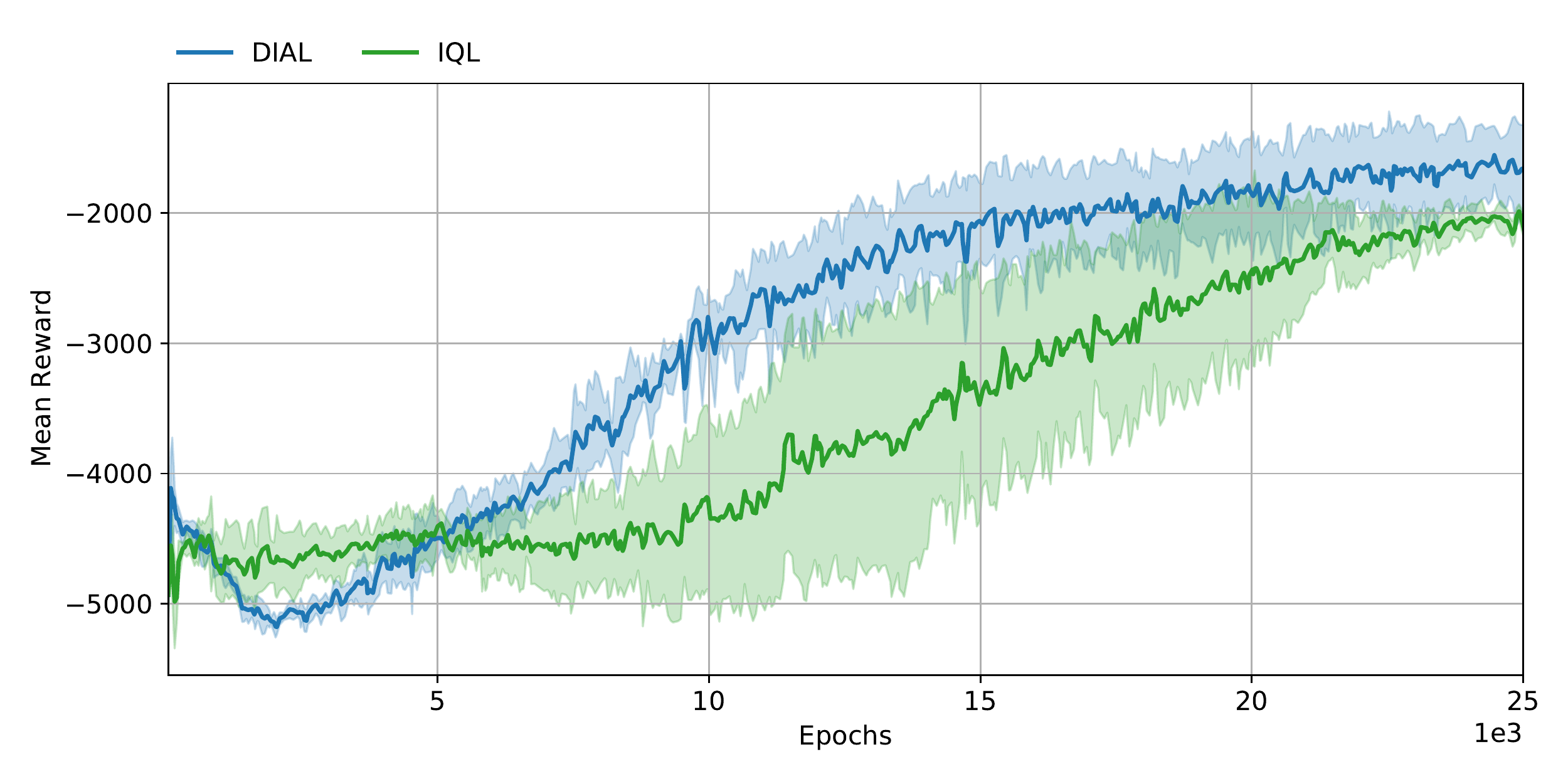}
\caption{The average training results of five runs for IQL and DIAL in the traffic light environment with the shaded area showing the standard deviation.}
\label{fig:training}
\end{figure}

The IQL and DIAL policies are trained using $25 * 10^3$ epochs and 35 million environment interactions.
% Define epoch
An epoch is defined as a training iteration in which we use 7 episodes with a fixed length of 200 timesteps. The agents are trained using a random inflow probability between 0.1\% and 60\%.
Every configuration is trained five times using a different random seed.
% Results
The average reward from these experiments are presented in Figure \ref{fig:training}. At epoch $5 * 10^3$, DIAL starts to outperform IQL in average reward while having a smaller standard deviation. Between epoch $22 * 10^3$ and $25 * 10^3$, the standard deviation of IQL becomes smaller as the standard deviation of DIAL but the average reward of DIAL remains higher than the average reward of IQL. These results show that DIAL is outperforming IQL both in learning speed and in average reward.

% TODO the reward of IQL does not rise much after this

Table \ref{table:peak_performance} shows the average reward of the best performing training iteration for both IQL and DIAL. These results show that DIAL is able to achieve a higher peak reward (8\% improvement) and achieve this faster than IQL.

\begin{table}[h]
\centering
\begin{tabular}{c|c|c|c}
Method                & Peak Performance Epoch & Average Peak Reward    & Peak Performance Standard Deviation           \\ \hline
\rule{0pt}{2ex}
IQL                   & $25.175 * 10^3$                & $-1986$                  & $54$                   \\
DIAL                  & $24.75 * 10^3$                 & $-1555$                  & $223$  \\
\end{tabular}
\caption{The average reward of five runs from the best performing training iteration for the IQL and DIAL method.}
\label{table:peak_performance}
\end{table}

\subsection{Evaluation}
Once the policies have been trained, a comparison on the performance is conducted by evaluating them on environments with fixed inflows as opposed to the random inflows used during training. Each policy iterates over ten fixed inflows and will be evaluated 200 times per inflow. The inflows are distributed evenly between the range of inflows used during training. The purpose of this experiment is to more accurately evaluate the behaviour of the trained policies without the random factor present during learning.
An additional configuration is added to gain insight into the need for communication during evaluation. The additional configuration will be using DIAL with disabled communication which is achieved by sending the message to both agents that was generated by the communication policy when no vehicles were observed. This will indicate whether communication is used as a bias or if it is actually used during evaluation.
If the communicating policy performance significantly drops when the agents are not allowed to communicate, communication is used for cooperation.

\begin{figure}[h]
\sidecaption
\includegraphics[scale=.5]{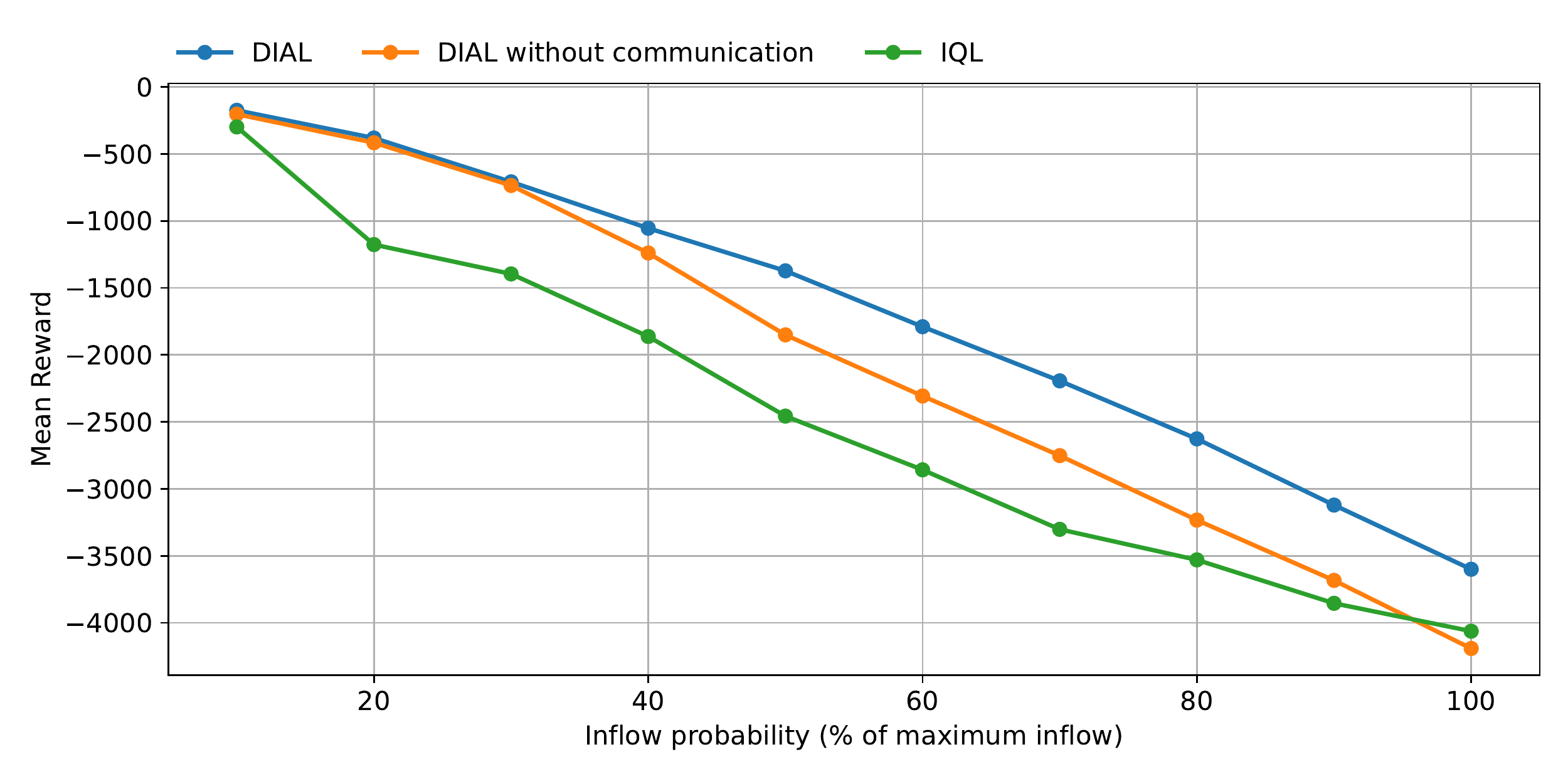}
\caption{Evaluation of the trained IQL and DIAL policies.}
\label{fig:evaluation}       % Give a unique label
\end{figure}

The result of this experiment can be found on figure \ref{fig:evaluation}.
First, it is clear that DIAL  outperforms IQL by a substantial amount. The difference is largest with average inflows ranging from 40\% to 80\%, slightly decreasing at the extremes. This indicates that the emergence of communication leads to a more optimal traffic control behaviour.
Finally, DIAL without communication is able to have similar performance for the low inflow probabilities but is outperformed by DIAL as the inflow probabilities increase.
Since the default message used in the DIAL without communication was generated when no vehicles are observed, this message will be close to the message that DIAL will send when there is a low inflow probability. Therefore DIAL without communication and DIAL will have a similar average reward for these low inflow probabilities.
DIAL without communication has a lower average reward than DIAL and IQL for an inflow probability of 100\% as the agents are receiving message from a low inflow probability situation.
This result shows that DIAL does not just have a better policy but has learned a valid communication protocol which increase the performance over a range of inflow probabilities.

%%%%%%%%%%%%%%%%%%%%%%%%%%%%%%%%%%%%%%%%%%%%%%%%%%%%%%%%%%%%%%%%%%%%%%%%%%%%%%%%%%%%%%%%%%%%%%%%%%%%%%%%%%%%%%%
% Conclusion
%%%%%%%%%%%%%%%%%%%%%%%%%%%%%%%%%%%%%%%%%%%%%%%%%%%%%%%%%%%%%%%%%%%%%%%%%%%%%%%%%%%%%%%%%%%%%%%%%%%%%%%%%%%%%%%
\section{Conclusion}
\label{sec:conclusion}
In this work, a multi-agent reinforcement learning agent for an adaptive traffic control system is investigated. We compare independent Q-learning with the DIAL method which has the ability to learn communication between the agents. These methods are compared on an environment with two traffic light, using the SUMO simulator, in which the agents can only observe a part of the entire state. In this configuration, we demonstrate that DIAL is able to learn faster and achieve a higher maximum reward than an independent Q-learner. These methods are evaluated on a range of inflow probabilities and the results demonstrate that DIAL is able to achieve a better result on all of the evaluated settings. We believe that these results show the importance of learning a communication policy simultaneously with the action policy for real world use cases in which not every agent has access to the full state but the agents need to learn how to share information in order to overcome this limited view of the current state.

% Future work
% More agents and use centralized critic
In future work, a more complex environment could be used that includes more agents that control more individual intersections in which the agent will learn to communicate with each other. This increase in number of agent will present certain challenges like the lazy agent problem (see Section \ref{sec:related_work}). This problem could be minimized by using a centralized critic which has been demonstrated by the MADDPG method \cite{lowe2017multi} and the MACC method \cite{vanneste2020learning}. Additionally, more communication topologies should be investigated. This will become especially important when the amount of agents increases and broadcasting every message becomes infeasible.

\begin{acknowledgement}
This work was supported by the Research Foundation Flanders (FWO) under Grant Number 1S94120N and Grant Number 1S12121N. We gratefully acknowledge the support of NVIDIA Corporation with the donation of the Titan Xp GPU used for this research.
\end{acknowledgement}
%
% \section*{Appendix}
% \addcontentsline{toc}{section}{Appendix}
% %
% %
% When placed at the end of a chapter or contribution (as opposed to at the end of the book), the numbering of tables, figures, and equations in the appendix section continues on from that in the main text. Hence please \textit{do not} use the \verb|appendix| command when writing an appendix at the end of your chapter or contribution. If there is only one the appendix is designated ``Appendix'', or ``Appendix 1'', or ``Appendix 2'', etc. if there is more than one.

% \begin{equation}
% a \times b = c
% \end{equation}

\bibliographystyle{spmpsci}
\bibliography{ref}

\end{document}